\documentclass[11pt,a4paper]{article}
\usepackage[hyperref]{paper}
\usepackage{times}
\usepackage{latexsym}

\usepackage{url}

\usepackage{graphicx}
\usepackage{booktabs}
\usepackage{mathtools} 
\usepackage{multirow} 
\usepackage{tabularx} 
\usepackage{subfig}
\usepackage{tablefootnote}
\usepackage{placeins}
\usepackage{amssymb}

\newcommand\blfootnote[1]{%
  \begingroup
  \renewcommand\thefootnote{}\footnote{#1}%
  \addtocounter{footnote}{-1}%
  \endgroup
}

\aclfinalcopy 

\newcolumntype{P}[1]{>{\centering\arraybackslash}p{#1}}

\title{Graph-Based Learning for Stock Movement Prediction with Textual and Relational Data}

\author{Qinkai Chen$^{\dag}$$^{\ddag}$ \, Christian-Yann Robert$^{\mathsection}$\\ 
$^{\dag}$Ecole Polytechnique, Palaiseau, France \\
$^{\ddag}$Exoduspoint Capital Management France, Paris, France \\
$^{\mathsection}$ENSAE Paris, Palaiseau, France \\
\tt{qinkai.chen@polytechnique.edu} \\
\tt{christian-yann.robert@ensae.fr}
}

\date{}

\begin{document}

\maketitle

\begin{abstract}
  Predicting stock prices from textual information is a challenging task due to the 
  uncertainty of the market and the difficulty in understanding the natural language from a machine's perspective. 
  Previous researches mostly focused on sentiment extraction based on single news. However,
  the stocks on the financial market can be highly correlated, one news regarding one stock can 
  quickly impact the prices of other stocks. To take this effect into account, we propose a 
  new stock movement prediction framework: Multi-Graph Recurrent Network for Stock Forecasting (MGRN). 
  This architecture allows to combine the textual sentiment from financial news 
  and multiple relational information extracted from other types of financial data. 
  Through an accuracy test and a trading simulation on the stocks of the STOXX Europe 600 index, 
  we demonstrate a better performance from our model than other benchmarks.
\end{abstract}

\section{Introduction}

\blfootnote{The authors would like to thank Mathieu Rosenbaum from Ecole Polytechnique for his valuable 
guidance and advice during this work.
The authors also appreciate the insightful discussions with Jean-Sebastien Deharo and Alexandre Davroux.}

\citet{fama1965behavior} and \citet{malkiel1989efficient} show that the movement of stock price can be
explained jointly by all known information, although it is volatile and 
non-stationary \citep{adam2016stock}.
The information can include all types of available information, such as historical prices \citep{kohara1997stock}, 
macroeconomic indicators \citep{garcia1999macroeconomic}, financial news \citep{ding2014using}, etc.
Most of the research focuses on the time series analysis of the numerical indicators, i.e., using historical 
prices to predict future prices \citep{luo2017improving}. Although simple and efficient, this method does not 
consider the market sentiment and market moving events, based on which most rational human investors trade.
With the development of the natural language processing, more recent research works start to use textual data for 
stock movement prediction \citep{ding2014using,ding2015deep,hu2018listening}. However, these researches assume 
that all the stocks are independent and predict the price movement of each stock independently, although \citet{hou2007industry}
shows that the movement of one stock can significantly impact other correlated stocks.

To take stock correlation into consideration, \citet{guo2018news} and \citet{ye2021multi} integrate the relationship 
information into traditional time series analysis without using textual data.  
\citet{cheng2020knowledge} and \citet{sawhney2020deep} design neural networks 
to take both textual data and one pre-defined relationship graph into consideration. However, the stock 
relationships can come from multiple aspects, such as price correlation \citep{campbell1993trading},
sector of activity \citep{vardharaj2007sector} and supply chain \citep{pandit2011information}. We 
will demonstrate that considering multiple relationships at the same time
can benefit the prediction performance.

Hence, we want to design an improved model which has the following characteristics: (1) 
learn from both text data and relational data, (2) incorporate an unlimited number of relational graphs
into the structure, (3) take temporal patterns of the news into account instead of learning from 
only one news at a time.

To address the above-mentioned challenges, we first discuss previous works (\textbf{Sec. 2}),
we then propose a new stock price movement prediction framework: 
Multi-Graph Recurrent Network for Stock Forecasting (MGRN). MGRN combines textual information 
from a financial news provider and relationship data from different sources to predict the variation
of stock prices (\textbf{Sec. 3}). MGRN jointly learns from texts and relationships through its 
graph-based structure, it can also learn from news' temporal patterns with its recurrent structure
(\textbf{Sec. 4}). With various experiments, we show 
the performance of our MGRN model as well as other benchmark models (\textbf{Sec. 5}). We also perform 
trading simulations to show the profitability of our results in real-life scenario (\textbf{Sec. 6}).

\section{Related Work}

\subsection{Stock Movement Prediction}

There are various approaches to predict stock prices and the researches on this topic span 
on different domains. Econometricians use time-series analysis \citep{mills1990time}
to predict future prices based on historical prices and volumes data.
Financial analysts rely on company fundamental data such as earnings and debt ratio 
\citep{ozlen2014effect,wang2004determines}, or macroeconomic data such as GDP and CPI index 
\citep{hoseinzade2019cnnpred} to predict the trend of stock prices from a economic point of view. 
Computer scientists tend to use machine learning techniques to interpret the stock price movement.
With the development of the natural language processing, more researches focus on
predicting stocks prices based on financial news or social media texts.

\citet{schumaker2009textual} use a classical feature engineering method to extract features from 
text data, \citet{ke2019predicting} use a TF-IDF \citep{crnic2011introduction} like method to 
identify positive and negative words in financial texts.
Nowadays, more researches adopt deep learning methods to analyze financial news.
\citet{ding2014using,ding2015deep} use structured representations and convolutional networks to
analyze news sentiments. \citet{hu2018listening} apply attention mechanism to directly handle
the raw text without using widely used recurrent neural network. \citet{xu2018stock} propose a model 
which considers jointly text and price information.
All these methods assume that all news are independent to simplify the problem. 
Although useful, this is contrary to the the common sense and some findings
\citep{hou2007industry,klossner2014exploring} which explain the price interactions among stocks.

\subsection{Graph Neural Network}

With the popularity of graph learning, more researchers start to use graph-based structure to capture
complex non-linear interactions among the nodes. 
Graph Convolutional Network (GCN) is one of the most used graph networks, 
and it has gained more popularity since it obtains outstanding result on node classification task
\citep{kipf2016semi}. Some recent researches apply this technique on stock movement prediction tasks.

\citet{chen2018incorporating} and \citet{kim2019hats} combine historical price and corporation 
relationship knowledge graph through graph-based models. However, they only take historical 
price data as input without considering the information from news or social media texts.
\citet{sawhney2020deep} design a Multipronged Attention Network (MAN-SF) to consider both 
textual data and relationship data at the same time. However, the study only considers one pre-built 
graph from Wikidata\footnote{https://www.wikidata.org/}. In the real world, the relationships among 
companies come from multiple dimensions and it can change significantly over time.

To close the gap in the researches, we propose MGRN, 
which can ingest both textual data and an unlimited number of 
relationship graphs built from different sources, as opposed to the previous researches. In addition,
MGRN contains a recurrent structure to model the temporal interactions of the news, 
instead of assuming the independence of the news.

\section{Problem Formulation}
\label{sec:problem_formulation}

Following \citet{ding2015deep} and \citet{xu2018stock}, we formulate the stock movement prediction as 
a binary classification task. Given a universe of stocks $S$, for a stock $s \in S$, 
we define its market adjusted return $r_{s}$ between $t$ and $t + \Delta t$ as:

\begin{equation}
  \label{eq:market_adjusted_return}
  r_{s, t} = \frac{P_{s, t + \Delta t}}{P_{s, t}} - \frac{P_{m, t + \Delta t}}{P_{m, t}}
\end{equation}

\noindent
where $P_{s, t}$ denotes the price for stock $s$ at time $t$, and $P_{m, t}$ denotes the market 
index value at time $t$.

We define the target of our stock movement prediction task for stock $s$ between $t$ and $t + \Delta t$ as:
\begin{equation}
  \label{eq:problem_formulation}
  Y_{s, t} =
  \begin{cases}
    1 , & r_{s, t} > 0\\
    0 , & r_{s, t} \leq 0\\
  \end{cases}
\end{equation}

For a traditional single stock movement prediction task, the goal is to predict $Y_{s,t}$ from all the news 
related to the stock $s$ in a look-back window $T$, it can be written as:
\begin{equation}
  \hat{Y}_{s, t} = f(E_{s, t}^{T}, \theta)
\end{equation}

\noindent
where $E_{s, t}^{T}$ denotes all the news for stock $s$ between $t-T$ and $t$ and $\theta$ denotes
the trainable parameters.

However, our goal is to consider both news and cross effects among stocks when predicting stock movement.
Our prediction is hence written as:
\begin{equation}
  \hat{Y}_{s, t} = f([E_{1, t}^{T},..., E_{n, t}^{T}], [G_{1},...,G_{g}], \theta)
\end{equation}

\noindent
where $n$ is the number of stocks in our universe $S$, $G_{i}$ is the graph constructed from data 
source $i$ and $g$ is the number of graphs we construct from different data sources.

\section{Multi-Graph Recurrent Network for Stock Forecasting}

\begin{figure*}[t]
  \includegraphics[width=\linewidth]{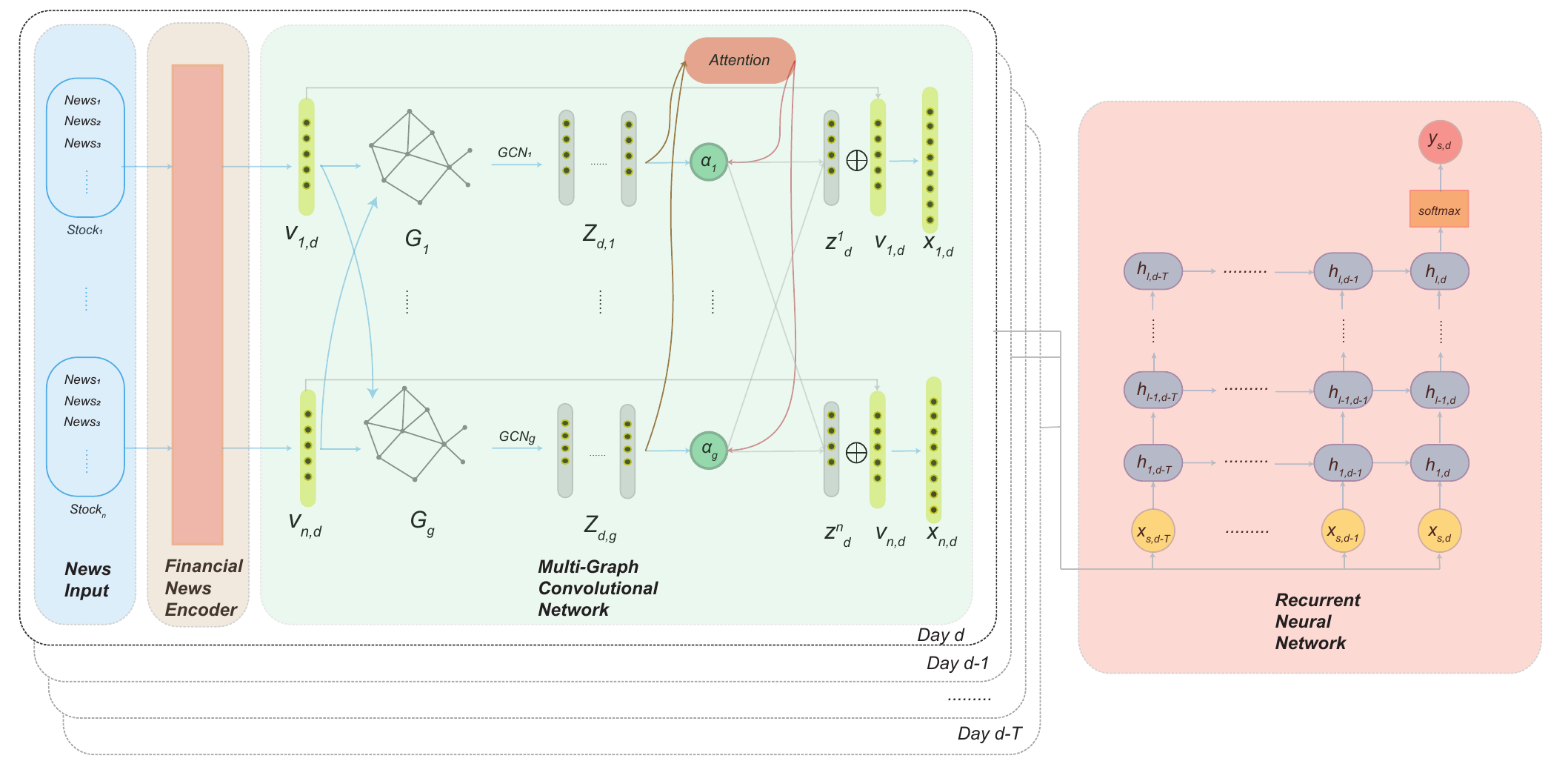}
  \caption{An overview of the architecture of the MGRN model.
  Our MGRN model includes three sub-components: (1) Financial News Encoder, which encodes textual news 
  into a fixed length vector for each stock and each day ($v_{s,d}$). 
  (2) Multi-Graph Convolutional Network, which takes the encoded daily news vectors and the graphs as 
  input. Through this multi-graph structure, we get multiple node embeddings for each stock. We then 
  combine these node embeddings into a single embedding ($\hat{x}_{s,d}$) through an attention mechanism.
  (3) Recurrent Neural Network, which takes the combined embeddings during a look-back window $T$ as 
  input and extracts temporal patterns among the news. $h_{i, j}$ denotes the \textit{j-th} LSTM cell on the \textit{i-th} layer.
  Finally, through a fully-connected layer, 
  we predict whether the stock price increases or decreases ($\hat{y}_{s,d}$).}
  \label{fig:model_graph}
\end{figure*}

The architecture of our MGRN model is shown in Figure \ref{fig:model_graph}.
It has three sub-components: Financial News Encoder, Multi-Graph Convolutional Network
and Recurrent Neural Network. We introduce the details of each component in the following subsections.

\subsection{Financial News Encoder}

\textbf{Single news embedding}

For each news $e$, we need to represent it with an embedding 
$v_{e} \in \mathbb{R}^{d}$. Following the work of \citet{sawhney2020deep}, we simply use 
Universal Sentence Encoder \citep{cer2018universal} to convert a sentence into a fixed-length 
embedding.

\bigskip
\noindent
\textbf{Aggregated news embedding} 

Unlike stock movement prediction based on single news, graph-based network structure requires a
valid node embedding for each node when we train and predict. Hence, we need to choose
a reasonable time window to make sure that for most of the stocks, there is at least one piece of news
in this window. This is to avoid too many zero vectors as node embeddings. We simply choose a
period of one day when we aggregate the news, following \citet{kim2019hats} and \citet{li2020modeling}.
It means that for stock $s$ and on day $d$, 
we select all the news concerning $s$ between the market close time 
on day $d$ and the market close time on day $d-1$ to get its aggregated embedding.

\citet{iyyer-etal-2015-deep} and \citet{wieting2015towards} show that a simple average aggregation
can have similar and even better performance than more complicated recurrent models such as 
LSTM. For the sake of simplicity without sacrificing the accuracy, we use an average over all 
news embeddings of a stock $s$ as its aggregated news embedding on day $d$. We denote it by $v_{s,d}$.
We have:
\begin{equation}
  \label{eq:daily_news_vector}
    v_{s,d} = \frac{1}{n}\sum^{|E_{s,d}^{1}|}_{i=1}e_{s,t}^{i}
\end{equation}

\noindent
where $e_{s,t}^{i} \in E_{s,d}^{1}$ is the embedding of 
the $i$-th news about $s$ happening at time $t$ between $d$ and $d-1$.

\subsection{Multi-GCN Attention Network}
\label{subsec:multi_gcn_attention_network}
\textbf{Graph Representation}

We model the stock relationships with a graph $G$. We use the graph's adjacency matrix 
$A \in \mathbb{R}^{n \times n}$ to represent the relationships among $n$ stocks. 
The element $A_{i, j}$ denotes the intensity of relationship between the stock $i$ and the stock $j$.
We set $A_{i, i} = 1$.

There are two types of relationships: (1) boolean relationship represented by a simple graph
and (2) continuous relationship represented by a weighted graph.

For a boolean relationship, we have $A_{i, j} \in \{0, 1\}$. If there is a connection between stock $i$
and $j$, $A_{i, j}$ is set to 1. Otherwise, it is set to 0.
For example, GICS sector \footnote{https://www.msci.com/gics} relationship is a boolean relationship. 
If two stocks are both in the same sector, we assert that they are connected. 
Supply chain relationship is also a boolean relationship. 
If one company is another company's supplier, we assert that they are connected.

However, for a continuous relationship, we have $A_{i, j} \in [0, 1]$. 
The more important the relation between two stocks, the larger this value.
For example, the historical price relationship is a continuous 
relationship. The intensity of the relationship between two stocks is calculated as the correlation coefficient of
two stocks' daily return time series.

Following \citet{duvenaud2015convolutional} and \citet{kipf2016semi}, we normalize our adjacency
matrix with a symmetric normalization:
\begin{equation}
  \hat{A} = D^{-\frac{1}{2}}AD^{-\frac{1}{2}}
\end{equation}
where $D \in \mathbb{R}^{n \times n}$ is a generalized diagonal node degree matrix for both simple graphs
and weighted graphs, defined as:

\begin{equation}
  D_{i, j} =
  \begin{cases}
    \sum_{k}A_{i,k} , & i = j \\
    0 , & i \neq j \\
  \end{cases}
\end{equation}

Such normalization guarantees that the operations involving $A$ do not change 
the scale of the result on both simple graphs and weighted graphs.

\bigskip
\noindent
\textbf{Single Graph Convolutional Network}

We use the same GCN structure as proposed by \citet{kipf2016semi}. For day $d$,
we construct our daily news matrix with $X_{d} = [v_{1,d}, ..., v_{n,d}]^T$.
We also have one graph $G$ and its adjacency matrix is $A$.

Our GCN with $L$ layers can be written as the following function:
\begin{equation}
  \label{eq:gcn}
  H^{(l+1)} = \sigma (\hat{A}H^{(l)}W^{(l)})
\end{equation}

\noindent
with $H^{(0)} = X_{d}$ and $H^{(L)} = Z_{d}$ as the final graph output.
We have $H^{(l)} \in \mathbb{R}^{n \times f_{l}}$ where $f_{l}$ denotes the number of output features
for layer $l$.
In Equation (\ref{eq:gcn}), $\sigma$ denotes the activation
function and $W^{(l)}$ represents the weight matrix for the layer $l$.

With such an operation, we obtain a new node representation of dimension $f_{L}$ 
for each stock from $H^{(L)}$.

\bigskip
\noindent
\textbf{Attention Aggregation Layer}

Given $g$ graphs $G_{1}, ... , G_{g}$ with their adjacency matrix $A_{1}, ... , A_{g}$,
we attribute each graph an independent GCN. For day $d$, we have $g$ graph outputs $Z_{d,1}, ..., Z_{d,g}$.
We combine these graph outputs to get an aggregated graph output with 
an attention mechanism \citep{vaswani2017attention}.

We define $W_{a} \in \mathbb{R}^{f_{L} \times w}$ and $q \in \mathbb{R}^{w \times 1}$,
both of which are trainable parameters.
We then calculate the attention coefficients $\alpha_{i} \in \mathbb{R}^{n \times 1}$
for graph $i$ using the following formula:
\begin{equation}
  \alpha_{i} = \frac{exp(Z_{d, i}W_{a}q)}{\sum_{j}exp(Z_{d, j}W_{a}q)}
\end{equation}

We then aggregate all the $Z_{d, i}$ using:
\begin{equation}
  Z_{d} = \sum_{i} \alpha_{i} \otimes Z_{d,i}
\end{equation}

\noindent
where $\otimes$ denotes element-wise multiplication.

Finally, we concatenate the graph output $Z_{d}$ with the original daily news embeddings.
Our final output after the graph layer for the day $d$ becomes:
\begin{equation}
  \hat{X_{d}} = X_{d} \oplus Z_{d}
\end{equation}
where $\oplus$ denotes concatenation. This is to ensure that we can capture the information
from both graphs and the orignal text embeddings.

\subsection{Recurrent Neural Network}
\label{subsec:recurrent_neural_network}


We then build a recurrent network to capture the temporal patterns in the news.

We first repeat the same process described in Section \ref{subsec:multi_gcn_attention_network} from 
day $d$ to day $d-T$. We have the outputs from the graph layer denoted by $\hat{X_{d}}, ..., \hat{X}_{d-T}$
as the input of our recurrent network.

We use a straightforward multi-layer recurrent neural network with 
LSTM cells \citep{hochreiter1997long} shown on the right-hand side of Figure \ref{fig:model_graph}.
At the final layer, we use a fully connected layer followed by a softmax to make the final prediction.

We input the concatenated outputs from the graph layer and financial news encoder layer sequentially
into the first layer of the RNN model. For each stock at each day, we get its $P^{+}_{s,d}$ denoting
the probability that the stock price will increase the next day and $P^{-}_{s,d} = 1 - P^{+}_{s,d}$
representing the price drop probability.

We train our MGRN network with an Adam optimizer \citep{kingma2014adam}
by minimizing the binary cross entropy loss, given as:
\begin{equation}
  l = \sum_{s} \sum_{d} Y_{s, d}ln(P^{+}_{s, d}) + (1 - Y_{s, d})ln(1 - P^{+}_{s,d})
\end{equation}

\noindent
where $Y_{s, d}$ is the true stock price movement defined in Equation \ref{eq:problem_formulation}.

\section{Experiments}

\subsection{Datasets and Graph Building}
\label{subsec:datasets}

\noindent
\textbf{Financial News Dataset}

The dataset that we use is Bloomberg News\footnote{https://www.bloomberg.com/professional/product/event-driven-feeds/}.
In this dataset, each entry contains a \textit{timestamp} denoting when this news happened, 
a \textit{ticker} which signifies the stock related to this news and the \textit{headline} of this news. 
In addition to the necessary information above, we can also find a \textit{score} which is among -1, 0 and +1, 
and a \textit{confidence} between 0 and 100 associated with the \textit{score}. 
These two fields are given by Bloomberg's proprietary classification algorithm, 
it will serve as one of the benchmarks for our prediction model. 
We present a sample dataset in Table \ref{tab:data_sample}.

\begin{table*}[!t]
  \centering
    \begin{tabular}{P{16em}P{7em}P{4em}cc}
    \toprule
    \textbf{Headline} & \textbf{TimeStamp} & \textbf{Ticker} & \textbf{Score} & \textbf{Confidence} \\
    \midrule
    1st Source Corp: 06/20/2015 - 1st Source announces the promotion of Kim Richardson in St. Joseph & 2015-06-20T05:02:04.063 & SRCE & -1 & 39 \\
    \midrule
    Siasat Daily: Microsoft continues rebranding of Nokia Priority stores in India opens one in Chennai & 2015-06-20T05:14:01.096 & MSFT & 1 & 98 \\
    \midrule
    Rosneft, Eurochem to cooperate on monetization at east urengoy & 2015-06-20T08:01:53.625 & ROSN RM & 0 & 98 \\
    \bottomrule
    \end{tabular}%
    \caption{A sample dataset from Bloomberg News dataset that we use as our financial news data.}
  \label{tab:data_sample}%
\end{table*}%

It is worth noting that we remove the stocks which do not have enough news.
This is to ensure that we do not have too many zero vectors as our daily news vector 
(Equation \ref{eq:daily_news_vector}).
We only select the stocks which have more than 2 news per day in average.
With a such filter, we have 168 stocks in the stock universe, and we observe that there are 
only 15\% (Table \ref{tab:data_stats}: Zero vector rate) 
zero vectors among all daily news vectors, meaning that given a stock and a date, there is 
a 85\% chance there is at least one piece of news.

\bigskip
\noindent
\textbf{Stock Price Dataset}

We extract all the market close prices for all the stocks in the universe, we also extract the 
Europe STOXX 600 index value at the market close time\footnote{17:30 Central European Time}
for our market adjusted return calculation.
We use the stock prices for both labelling and building a correlation graph from stock returns.

For labelling, we follow the procedure described in Section \ref{sec:problem_formulation}. 
However, we observe that there are some delisted 
stocks which no longer have prices after a certain date, preventing us from correctly
calculating their returns. Hence, we remove the stocks which are delisted during our training
period. There are three such stocks, leaving us 165 stocks in total in our experiments.

We also use the stocks prices to build a weighted graph $G_{c}$.
For all stocks, we first calculate its market adjusted returns with Equation \ref{eq:market_adjusted_return},
we have a vector $v_{s} = [r_{s, 1}, ..., r_{s, T_{c}}]$ containing all the returns from the 
first day until the last day in our training dataset. We calculate the Pearson Correlation Coefficient
\citep{freedman2007statistics}
between stock $i$ and stock $j$, such that its adjacency matrix $A_{c}$ is given by:

\begin{equation}
  \label{eq:correlation_graph}
  A_{c, i, j} = \frac{cov(v_{i}, v_{j})}{std(v_{i})std(v_{j})}
\end{equation}

\noindent
where $cov$ represents the covariance and $std$ denotes the standard deviation.

\bigskip
\noindent
\textbf{Stock Sector Data}

In finance, each company is classified into a specific sector with Global Industry Classification Standard
(GICS). We use this data to construct a sector graph $G_{s}$. Its adjacency matrix $A_{s}$ is defined as:
\begin{equation}
  \label{eq:sector_graph}
  A_{s, i, j} = 
  \begin{cases}
    1 , & sector(i) = sector(j) \\
    0 , & otherwise \\
  \end{cases}
\end{equation}

There are four granularities in GICS sector data: Sector, Industry Group, Industry, Sub-Industry.
We can therefore construct four graphs with this dataset. In our experiments, we use the Industry
granularity as it gives the best performance. The performances with different sector graphs 
are discussed in Section \ref{subsec:experiment_results}

\bigskip
\noindent
\textbf{Supply Chain Data}

We use the supply chain data from 
Factset\footnote{https://www.factset.com/marketplace/catalog/product/factset-supply-chain-relationships}
to construct a supply chain graph.
This dataset describes the supplier-customer relationship (SCR) among different companies.
We construct a supply chain graph $G_{sc}$ such that
\begin{equation}
  \label{eq:supply_chain_graph}
  A_{sc, i, j} = 
  \begin{cases}
    1 , & i\ and\ j\ have\ SCR  \\
    0 , & otherwise \\
  \end{cases}
\end{equation}

\begin{figure*}
  \includegraphics[width=\linewidth]{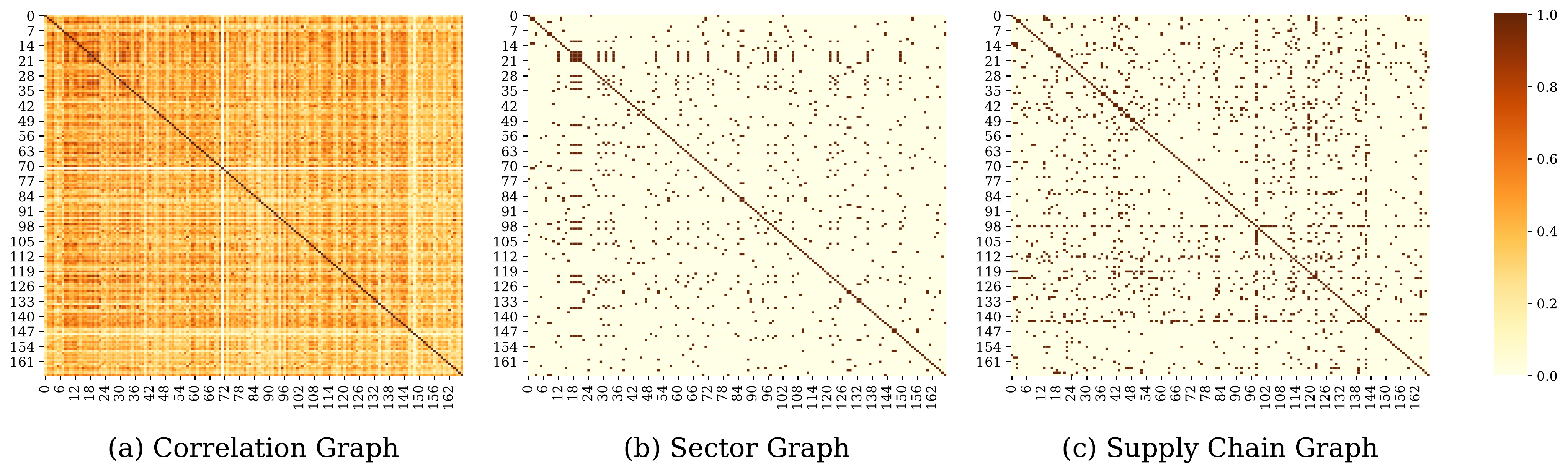}
  \caption{The heatmaps of our three graphs $G_{c}$, $G_{s}$ and $G_{sc}$.
  We can see some common characteristics in these heatmaps, for example,
  the top-left corner of the correlation graph and the sector graph.
  However, the graphs are rather uncorrelated, we prove this with
  the experiment results in Section \ref{subsec:experiment_results}.}
  \label{fig:heatmaps}
\end{figure*}

We show the heatmaps of three graphs in Figure \ref{fig:heatmaps}.

\bigskip
\noindent
\textbf{Dataset Split}

Following the standard in deep learning researches, we split our dataset into three sub-datasets:
train, dev and test. The details are shown in Table \ref{tab:data_stats}.

\begin{table}[h!]
  \centering
  \small
    \begin{tabular}{cccP{5em}}
    \toprule
      & \textbf{Train} & \textbf{Dev} & \textbf{Test} \\
    \midrule
    Total news & 1,199,367 & 316,944 & 439,949 \\
    Start & 01/2016 & 07/2018 & 01/2019 \\
    End & 06/2018 & 12/2018 & 12/2019 \\
    Nb. Stocks & 165 & 165 & 165 \\
    Trading days & 652 & 118 & 256 \\
    Data points\tablefootnote{Number of stocks multiplied by number of trading days,
    this equals the total predictions we make in each dataset.} 
    & 107,580 & 19,470 & 42,240 \\
    Zero vector rate & 15\% & 17\% & 19\% \\
    \bottomrule
    \end{tabular}%
    \caption{Statistics of the news dataset. Zero vector rate means the ratio of zero vector among
    all embedded daily news vectors $v_{s, t}$. We only select the 165 stocks which have relatively
    more news to make this value as small as possible in order not to impact our GCN model.}
  \label{tab:data_stats}%
\end{table}%

\bigskip
\noindent
\textbf{Parameter Settings}

We use a look-back window $T=20$ days and we use a look-forward window $\Delta t=1$ day
to label our data.

The GCN model we use has two hidden layers, with 128 and 64 dimensions respectively. Our RNN
model also has two layers, with 128 and 64 LSTM cells respectively. We train our model
with an Adam optimizer for 10 epochs. We set the batch size to 32.

\subsection{Evaluation Metrics}
\label{subsec:evaluation_metrics}

\textbf{Accuracy}

Following previous researches on the stock movement prediction task
\citep{ding2015deep,hu2018listening,sawhney2020deep}, we use accuracy 
to evaluate the performance of our model.

However, this simple metric does not reflect the need of a real-life investor,
since he does not need to make trades on all prediction results. The investor only trades when he is 
more confident about the prediction. In other words, the accuracy on the predictions with 
higher probability is more important than those with a mediocre probability.
Hence, we also include "percentile accuracy" in our evaluation metrics.

We note that the score $S_{s, d} \in [-1,1]$ for a stock $s$ on day $d$ as:
\begin{equation}
  S_{s, d} = (P^{+}_{s, d} - 0.5) \times 2
\end{equation}

For \textbf{each day}, we choose the top $\frac{q}{2}$-percentile scores and 
the bottom $\frac{q}{2}$-percentile scores of that day, where $q$ is a value between 0 and 100. 
We denote the accuracy calculated based on such selection as $Acc_{q}$. 
The accuracy on the whole test set is therefore $Acc_{100}$.

\bigskip
\noindent
\textbf{Trading Simulation}

We use a simple long/short trading strategy similar to \citet{ke2019predicting}.
For each day, we attribute equally weighted long positions for the stocks 
whose scores are in the top $\frac{q}{2}$-percentile.
For the stocks whose scores are in the bottom $\frac{q}{2}$-percentile, 
we give each stock the same short position. In this case, our long position equals to 
our short position, leaving no market exposure for our strategy.

We use annualized return and Sharpe ratio \citep{sharpe1994sharpe} to evaluate the performance of our strategies.
The annualized Sharpe ratio is defined as the ratio of the expected return $R$ to its standard deviation
multiplied by square root of the number of trading days $D_{y}$ in one year:
\begin{equation}
  Sharpe = \frac{\mathbf{E}(R)}{\sigma(R)} \times \sqrt{D_{y}}
\end{equation}

\subsection{Baseline Models}

\begin{table*}[t]
  \centering
    \begin{tabular}{ccccccccccc}
    \toprule
    $q$     & \multicolumn{2}{c}{100} & \multicolumn{2}{c}{50} & \multicolumn{2}{c}{20} & \multicolumn{2}{c}{10} & \multicolumn{2}{c}{2} \\
    \midrule
      metric  & Ret.\tablefootnote{The annualized return is shown in \%} 
       & Sharpe & Ret.  & Sharpe & Ret.  & Sharpe & Ret.  & Sharpe & Ret.  & Sharpe \\
    \midrule
    RAND  & 0.53  & 0.25  & 0.29  & 0.05  & -0.5  & -0.12 & -0.6  & 0.11  & 3.59  & 0.33 \\
    ARIMA & 0.3   & 0.09  & 0.79  & 0.16  & 2.26  & 0.9   & 1.5   & 0.13  & 2.24  & 0.36 \\
    BBG   & \textbf{2.78} & 0.66  & \textbf{3.72} & 0.57  & 3.89  & 0.56  & 2.56  & 0.35  & 18.48 & \textbf{1.54} \\
    Mean-BERT & 0.21 & 0.05 & 0.98 & 0.41 & 4.25 & 0.66 & 7.21 & 0.98 & 8.08 & 1.11 \\
    MAN-SF & 0.17 & 0.58 & 0.41 & 0.13 & 1.06 & 0.32 & 3.77 & 0.57 & 4.02 & 0.37 \\
    \midrule
    RNN   & 0.74  & 0.31  & 1.01  & 0.3   & 3.09  & 0.57  & 4.36  & 0.67  & 5.32  & 0.9 \\
    MGRN-Corr & 1.27  & 0.4   & 2.04  & 0.51  & 3.45  & 0.57  & 5.06  & 0.61  & 15.19 & 1.19 \\
    MGRN-Sector & 1.22  & 0.39  & 2.47  & 0.51  & 3.67  & 0.62  & 5.26  & 0.79  & 8.42  & 0.57 \\
    MGRN-Supply & 1.05  & 0.42  & 1.92  & 0.58  & 3.11  & 0.72  & 10.86 & 1.31  & 10.55 & 0.7 \\
    MGRN  & 2.18  & \textbf{0.94} & 2.07  & \textbf{0.62} & \textbf{8.71} & \textbf{1.7} & \textbf{12.03} & \textbf{1.33} & \textbf{26.22} & 1.51 \\
    \bottomrule
    \end{tabular}%
    \caption{The trading simulation result of all models with different $q$-percentiles.}
  \label{tab:res_backtest}%
\end{table*}%

We compare the performance of our MGRN model with other baseline models to demonstrate its performance.

We include the following baseline models:
\begin{itemize}
  \item {\tt RAND}: Random guess of $Y_{s, t}$.
  \item {\tt ARIMA}: Auto-Regressive Integrated Moving Average model \citep{ho1998use}
   based on historical prices.
  \item {\tt BBG}: The prediction given by Bloomberg which comes along with Bloomberg News dataset (Table \ref{tab:data_sample}).
  \item {\tt Mean-BERT}: We fine-tune the Bidirectional Encoder Representations from Transformers (BERT) model proposed by \citet{devlin2018bert}
  as a classification model. We use the average score of all the news for stock $s$ on day $t$ as its $S_{s,d}$.
  \item {\tt MAN-SF\footnote{MAN-SF only allows to have one relationship, we use the correlation for this model.}}: 
  A stock movement prediction framework proposed by \citet{sawhney2020deep}. The model combines price data, news data
  and relational data to predict stock return.
  \item {\tt RNN}: The model introduced in Sec. \ref{subsec:recurrent_neural_network} without adding 
   any graph. This is the same as a MGRN model with an identity matrix as graph adjacency matrix.
\end{itemize}

To make a detailed analysis of the improvement brought by different graphs, 
we train our MGRN model with different graphs:
\begin{itemize}
  \item {\tt MGRN-Corr}: MGRN model with return correlation graph $G_{c}$ (Eq. \ref{eq:correlation_graph}).
  \item {\tt MGRN-Sector}: MGRN model with sector graph $G_{s}$ (Eq. \ref{eq:sector_graph}).
  \item {\tt MGRN-Supply}: MGRN model with supply chain graph $G_{sc}$ (Eq. \ref{eq:supply_chain_graph}).
  \item {\tt MGRN}: the full MGRN model using three graphs $G_{c}$, $G_{s}$ and $G_{sc}$ at the same time.
\end{itemize}

\subsection{Experiment Results}
\label{subsec:experiment_results}

Table \ref{tab:res_accuracy} shows the accuracy of different models on the test set
with different $q$-percentiles. We find that our MGRN model shows the best performance,
outperforming other baseline models.

We compare the single graph models (MGRN-Corr, MGRN-Sector and MGRN-Supply) and the 
vanilla model without graph (RNN). We find that all the graphs can help improve 
the performance, especially for the most extreme scores (a smaller $q$ value).
However, it is difficult to say which graph has the best performance, since each
graph has different optimal performances on different percentiles.
For example, the supply chain graph has the most added value on the most extreme 
scores (highest with $q=2$), while the return correlation graph is more powerful 
on less extreme scores (highest with $q=10$ and $q=20$). This also signifies
that the information in each graph is rather independent, making it more reasonable
to combine different graphs.

We validate our hypothesis that combining different graph can help improve model performance
by comparing the full model (MGRN) with the single graph models. We find that 
when using all three graphs together, we have a significant improvement in accuracy
(5\% with $q=10$ and 3.5\% with $q=20$).
It proves that our model can absorb necessary information from multiple independent graphs 
at the same time, validating the effectiveness of combining relationship information
from different sources.

We also notice that adding a graph can lead to a worse result compared with the no-graph RNN in some scenario,
for example, MGRN-Supply is worse than RNN when $q=10$ and $q=20$. However, when combining with other graphs,
the result is better than using any graph individually.
This is because the errors usually come from several particular stocks, especially when we only have
only one source of information. If the source is incorrect, it can lead to significant error.
The benefit of using multiple graphs is to reduce the impact of these cases by making decisions
based on more than one source of information.

Table \ref{tab:res_backtest} shows the trading simulation result using the strategy described
in Sec. \ref{subsec:evaluation_metrics}. We can also confirm that our MGRN model
outperforms other models and that combining the graphs together is beneficial.
Although sometimes Bloomberg Sentiment Score shows 
better stability (Sharpe Ratio), MGRN model is still the model that consistently gives the best performance.
This validates the usage of MGRN model in real-world scenario.

\begin{table}[h]
  \small
  \centering
    \begin{tabular}{cccccc}
    \toprule
    $q$     & 100   & 50    & 20    & 10    & 2 \\
    \midrule
    RAND  & 0.471 & 0.471 & 0.472 & 0.473 & 0.488 \\
    ARIMA & 0.479 & 0.509 & 0.521 & 0.512 & 0.519 \\
    BBG\tablefootnote{As the Bloomberg Sentiment Score is a three class classification, 
    we remove all the neutral predictions to be comparable with our two class classification
    result} 
    & 0.501 & 0.500 & 0.487 & 0.488 & 0.551 \\
    Mean-BERT & 0.518 & 0.528 & 0.561 & 0.593 & 0.665 \\
    MAN-SF & 0.504 & 0.499 & 0.516 & 0.530 & 0.599 \\
    \midrule
    RNN   & 0.515 & 0.521 & 0.545 & 0.580 & 0.690 \\
    MGRN-Corr & 0.516 & 0.531 & 0.576 & 0.623 & 0.696 \\
    MGRN-Sector & 0.515 & 0.524 & 0.550 & 0.580 & 0.709 \\
    MGRN-Supply & 0.515 & 0.522 & 0.534 & 0.557 & 0.720 \\
    MGRN  & \textbf{0.522} & \textbf{0.537} & \textbf{0.580} & \textbf{0.633} & \textbf{0.740} \\
    \bottomrule
    \end{tabular}%
    \caption{The accuracy of baseline models and MGRN models with different $q$-percentiles.}
  \label{tab:res_accuracy}%
\end{table}%

\noindent
\textbf{Sector Graphs}

As we mentioned in Section \ref{subsec:datasets}, there are four granularities in our
GICS sector data. We compare the performances from all four granularities, and we find that 
the Industry level (the third granularity) shows the best performance,
especially on more extreme scores. 
Hence, we choose to use Industry level to build $G_{s}$. 
The detailed result is shown in Table \ref{tab:res_sector}.

\begin{table}[htbp]
  \centering
  \small
    \begin{tabular}{ccccc}
    \toprule
    \textbf{level} & \textbf{name}  & $q$=100   & $q$=20    & $q$=10 \\
    \midrule
    1     & Sector & 0.519 & 0.521 & 0.542 \\
    2     & Industry Group & 0.514 & 0.529 & 0.569 \\
    3     & Industry & 0.515 & \textbf{0.550} & \textbf{0.580} \\
    4     & Sub-Industry & 0.509 & 0.542 & 0.556 \\
    \bottomrule
    \end{tabular}%
    \caption{The accuracy of MGRN-Sector model but with the sector graphs 
    built from different GICS sector granularities.}
  \label{tab:res_sector}%
\end{table}%

\subsection{Qualitative Analysis: An Example}

We give a detailed study on one specific case to show how our MGRN model helps improve stock movement prediction.

We focus on the stock \textbf{TLW LN}\footnote{Tullow Oil plc is a multinational oil and gas exploration company.}
on the Dec. 6, 2018. We notice a news in the evening of that day: \textit{Tullow Oil Chairman Thompson Acquires Shares}.
This is a positive signal since the executive of Tullow Oil buys its shares, showing confidence as an insider. Based on this piece of news among others, our vanilla MGRN (RNN)
without any relational input gives a slightly positive score for this stock at 0.025. However, we observe a return of -7.7\% on the next trading day
which is contrary to our prediction result.

If we look at the same prediction from MGRN-Sector model, we find that its $S_{s,d}$ equals -0.107, which is a correctly predicted negative value among the bottom 10-percentile.
The only reason this new prediction is very different from that of vanilla MGRN is the impact from other related stocks.
We find that \textbf{GLEN LN}\footnote{Glencore plc is an Anglo-Swiss multinational commodity trading and mining company.}
has the most negative score from vanilla MGRN in the same sector. When we look at the news, we can find plenty of negative news about this company on the same day,
such as \textit{Rosen Law Firm Announces Investigation of Securities Claims Against Glencore plc}. These negative news caused the price drop of \textbf{GLEN LN}
by 3.4\%, which potentially caused the negative return (-2.6\%) in the same sector since we do not observe many negative news about other companies.

We can also see the same phenomenon with MGRN-Corr since the correlation between two stocks are relatively high (0.56),
but the prediction from MGRN-Supply is still false because there is no supplier-customer relationship between these two stocks.
We show the detail of this analysis in Table \ref{tab:case_study}.

\begin{table}[htbp]
  \centering
  \small
  \begin{tabular}{ccccc}
    \toprule
    Model & $A_{i,j}$ & Ticker & Score & Result \\
    \midrule
    \multirow{2}[2]{*}{RNN} & \multirow{2}[2]{*}{0} & TLW LN & 0.025 & False \\
    &       & GLEN LN & -0.055 & True \\
    \midrule
    \multirow{2}[2]{*}{MGRN-Corr} & \multirow{2}[2]{*}{0.56} & TLW LN & -0.036 & \textbf{True} \\
    &       & GLEN LN & -0.031 & True \\
    \midrule
    \multirow{2}[2]{*}{MGRN-Sector} & \multirow{2}[2]{*}{1} & TLW LN & -0.107 & \textbf{True} \\
    &       & GLEN LN & -0.031 & True \\
    \midrule
    \multirow{2}[2]{*}{MGRN-Supply} & \multirow{2}[2]{*}{0} & TLW LN & 0.013 & False \\
    &       & GLEN LN & -0.055 & True \\
    \bottomrule
  \end{tabular}%
  \caption{Detailed results of the case study for \textbf{TLW LN} on the Dec. 6, 2018.
  MGRN-Corr and MGRN-Sector both give correct results because the negative signal from
  \textbf{GLEN LN} can reach \textbf{TLW LN} through the graphs, but MGRN-Supply still gives the wrong prediction
  since these two stocks do not have connection on this graph.}
  \label{tab:case_study}%
\end{table}%

This example shows clearly how our MGRN model helps improve prediction result compared with a traditional recurrent model without relational modelling:
the related stocks can transmit their information through the meaningful graph. The model can then make decision based on both its own information
and the transmitted information.

\section{Conclusion}
We predict the stock movement by jointly considering financial news, multiple graph-based features
and temporal patterns of the news.
We introduce Multi-Graph Recurrent Network (MGRN) for this task. Through extensive 
experiments and trading simulations, we demonstrate the effectiveness of the model structure.
The result also proves that adding relationship information, especially different 
relationship information from multiple sources, can help better predict stock movement.
We plan to incorporate more types of data (such as time series) in our model to 
further improve the prediction accuracy.

\section*{Acknowledgments}

The authors gratefully acknowledge the financial support of the Chaire 
\textit{Machine Learning \& Systematic Methods}
and the Chaire \textit{Analytics and Models for Regulation} of Ecole Polytechnique.



\bibliography{paper}
\bibliographystyle{paper}

\end{document}